\def\isarxiv{1}

\ifdefined\isarxiv
\def\paperTitle{Only Large Weights (And Not Skip Connections) \\Can Prevent the Perils of Rank Collapse}
\else
\def\paperTitle{Understanding the Expressive Power of Attention from a Rank Collapse Perspective}
\fi

\def\paperAuthor{
Josh Alman\thanks{\texttt{josh@cs.columbia.edu}. Columbia University.}
\and
Zhao Song\thanks{\texttt{magic.linuxkde@gmail.com}. University of California, Berkeley.}
}

\ifdefined\isarxiv
\documentclass[11pt]{article}
\usepackage[numbers]{natbib}
\else
\documentclass{article}
\usepackage{neurips_2025}
\fi

\ifdefined\isarxiv

\usepackage{amsmath}
\usepackage{amsthm}
\usepackage{amssymb}
\usepackage{algorithm}
\usepackage{subfig}
\usepackage{algpseudocode}
\usepackage{graphicx}
\usepackage{grffile}
\usepackage{wrapfig,epsfig}
\usepackage{url}
\usepackage{xcolor}
\usepackage{epstopdf}

\usepackage{bbm}
\usepackage{dsfont}

\else 

\usepackage[utf8]{inputenc} % allow utf-8 input
\usepackage[T1]{fontenc}    % use 8-bit T1 fonts
\usepackage{hyperref}       % hyperlinks
\usepackage{url}            % simple URL typesetting
\usepackage{booktabs}       % professional-quality tables
\usepackage{amsfonts}       % blackboard math symbols
\usepackage{nicefrac}       % compact symbols for 1/2, etc.
\usepackage{microtype}      % microtypography
\usepackage{xcolor}         % colors

\fi

\ifdefined\isarxiv

\else
\usepackage{amsmath}
\usepackage{amsthm}
\usepackage{amssymb}
\usepackage{algorithm}
\usepackage{subfig}
\usepackage{algpseudocode}
\usepackage{graphicx}
\usepackage{grffile}
\usepackage{wrapfig,epsfig}
\usepackage{epstopdf}
\usepackage{bbm}
\usepackage{dsfont}
\fi
 
\allowdisplaybreaks

\ifdefined\isarxiv

\usepackage{tikz}
\usepackage{hyperref}  
\hypersetup{colorlinks=true,citecolor=blue,linkcolor=blue} 
\usetikzlibrary{arrows}
\usepackage[margin=1in]{geometry}

\else
\definecolor{mydarkblue}{rgb}{0,0.08,0.45}
\hypersetup{colorlinks=true, citecolor=mydarkblue,linkcolor=mydarkblue}
\fi
 
\graphicspath{{./figs/}}

\theoremstyle{plain}
\newtheorem{theorem}{Theorem}[section]
\newtheorem{lemma}[theorem]{Lemma}
\newtheorem{definition}[theorem]{Definition}

\newtheorem{corollary}[theorem]{Corollary}

\newtheorem{fact}[theorem]{Fact}
\newtheorem{remark}[theorem]{Remark}

\newcommand{\wt}{\widetilde}

\newcommand{\R}{\mathbb{R}}

\DeclareMathOperator{\diag}{diag}

\DeclareMathOperator{\SAtt}{\mathsf{SAtt}}
\DeclareMathOperator{\Res}{\mathsf{Res}}
\DeclareMathOperator{\soft}{\mathsf{softm}}

\DeclareMathOperator{\softmv}{\mathsf{softmv}}

\begin{document}

\ifdefined\isarxiv
%%% The below part is the title and author of ArXiv version.

\date{}
\title{\paperTitle}
\author{\paperAuthor}

\else
%%% The below part is the title and author of NeurIPS version.

\title{\paperTitle}

\author{%
  David S.~Hippocampus\thanks{Use footnote for providing further information
    about author (webpage, alternative address)---\emph{not} for acknowledging
    funding agencies.} \\
  Department of Computer Science\\
  Cranberry-Lemon University\\
  Pittsburgh, PA 15213 \\
  \texttt{hippo@cs.cranberry-lemon.edu} \\
  % examples of more authors
  % \And
  % Coauthor \\
  % Affiliation \\
  % Address \\
  % \texttt{email} \\
  % \AND
  % Coauthor \\
  % Affiliation \\
  % Address \\
  % \texttt{email} \\
  % \And
  % Coauthor \\
  % Affiliation \\
  % Address \\
  % \texttt{email} \\
  % \And
  % Coauthor \\
  % Affiliation \\
  % Address \\
  % \texttt{email} \\
}

\maketitle

\fi

\ifdefined\isarxiv
\begin{titlepage}
  \maketitle
  \begin{abstract}

%%%Below is Zhao's first version
%Attention-based structure has been ubiqutius in nowdays machine learning. In this work, we explore the experssibility of attention unit in Large lagunage models. Previous work has been working on two possible directions about when LLM has reasonable expressibility, one is large weights, and the other is residual structure. In this work, we prove that if we hope the transformer can represent certain complicated structure, using residual is actually not going to help. Which means, only big weights will help.

%%%Above is Zhao's first version

\ifdefined\isarxiv
%Here's the good one for arxiv:

Attention mechanisms lie at the heart of modern large language models (LLMs). Straightforward algorithms for forward and backward (gradient) computation take quadratic time, and a line of work initiated by [Alman and Song NeurIPS 2023] and [Alman and Song NeurIPS 2024] has shown that quadratic time is necessary unless the model weights are small, in which case almost linear time algorithms are possible. In this paper, we show that large weights are necessary to avoid a strong preclusion to representational strength we call layer collapse, which means that the entire network can be approximated well by a network with only a single layer. Thus, the quadratic running time of attention is unavoidable for expressive transformers.

The notion of layer collapse that we introduce is a variant on the notion of rank collapse from the work of [Dong, Cordonnier, and Loukas ICML 2021]. They showed that in Self Attention Networks with small weights and with skip connections, rank collapse must occur. This is typically interpreted as justifying the necessity of skip connections in expressive networks. However, our result shows that even with skip connections, if the weights are small, then layer collapse still occurs. Thus, only large weights, and not skip connections, can prevent these representational weaknesses.

\else

Attention mechanisms lie at the heart of modern large language models (LLMs), enabling them to capture complex dependencies within sequence data. We investigate the expressive capacity of a single attention unit in transformers, contrasting a prevailing hypothesis: the expressiveness of Transformers is both rooted in parameter magnitude and architectural features like residual connections. Through in-depth theoretical analysis, we show that residual links alone do not substantially expand the class of representable functions of Transformers. Instead, it is the scale of the attention weights (i.e., the model’s parameter magnitude), that determines its ability to encode complex structures. This finding overturns the notion that deeper architectures with residual pathways suffice for high expressivity and instead identifies weight expansion as the primary vehicle for augmenting a transformer’s representational power. Our results offer concrete guidance for LLM design, prioritizing parameter scaling over residual topologies when modeling highly intricate relationships.

Our main result concerns layer collapse, a new notion that we introduce, which means that the model can be approximated very well by a single-layer transformer. We show that even with skip connections, transformers experience layer collapse unless they have large weights.

\fi

  \end{abstract}
  \thispagestyle{empty}
\end{titlepage}

{\hypersetup{linkcolor=black}
%\tableofcontents
}
\newpage

\else

\begin{abstract}

\end{abstract}

\fi

%%% The part below is the main body and reference

%%% This file is the structure for main body content
%%% This file should only contain use \input{xxx}.
%%% TeX files for body contents should be named as:
%%% 01_xxxx.tex
%%% 02_xxxx.tex
%%% ...
%%% 49_xxxx.tex

\section{Introduction}

The rapid progress of large language models, text-to-image and text-to-video models like 
Transformer \cite{vsp+17}, 
BERT \cite{dclt18}, 
GPT-4 \cite{o23}, Llama 3 \cite{llama3}, Gemini 2.0 \cite{gemini}, and Adobe Firefly 3 \cite{firefly3}, 
has enabled powerful language modelling abilities. These models take advantage of large-scale pretraining on massive textual data, which equips them with strong abilities to interpret the complex patterns of natural language. These LLMs have a broad range of applications, influencing domains such as human-computer interaction, multilingual translation, language comprehension, text generation, and rapid prototyping of software. 

The major architecture behind the success of all these language models is the attention mechanism. Specifically, attention computes pairwise similarities by calculating inner products between vectorized representations of words, with input sequences represented as vectors. Formally, softmax attention can be formulated as follows:
\begin{definition}[Self-Attention with Softmax Units]\label{def:SAtt}
     Let $A \in \R^{n \times d}$ and weights $Q,K, V \in \R^{d \times d}$. Let $g$ represent the entry-wise exponentiation function, i.e., for $z\in \R$ we have $g(z) = \exp(z)$, and for a matrix $W$ we have $g(W)_{i, j} = g(W_{i, j})$. The attention computation can be defined as 
\begin{align*}
    \mathsf{SAtt}(X,Q,K,V) = \underbrace{ D^{-1} }_{n \times n} \underbrace{ g( X QK^\top X^\top ) }_{n \times n} \underbrace{ X }_{n \times d} \underbrace{ V }_{d \times d}
\end{align*}
where $D:= \diag( g( X Q K^\top X^\top ) {\bf 1}_n )$, and where ${\bf 1}_n \in \R^n$ is a length-$n$ vector whose entries are all $1$.
\end{definition}

\paragraph{Small Coefficients are Needed for Fast Algorithms}
However, the straightforward algorithm for computing self-attention results in a quadratic $O(n^2d)$ running time, where $n$ is the length of the input token and $d$ is the hidden dimension. Under popular complexity-theoretic assumptions, there is no better, subquadratic time algorithm to compute attention, even approximately \cite{as23}. Therefore, models based on attention may face difficulties when they handle long contexts.

In fact, a key observation of this line of work on the computational complexity of attention is that attention can be computed (or tightly approximated) faster if one restricts to small weights, i.e., an upper bound on how large the entries of $Q,K,V$ can be in Definition~\ref{def:SAtt} above. Indeed, a line of work \cite{as23,as24_iclr,as24_neurips,as25} has shown that small weights are both necessary and sufficient for a faster algorithm: If the weights are large, then the aforementioned complexity-theoretic result shows that there is no subquadratic time algorithm. However, if the weights are small, then attention can be approximated to low error in \emph{almost linear time}! Their algorithm is based on low-rank approximations of the $n \times n$ attention matrix (the matrix $g( X QK^\top X^\top )$ in Definition~\ref{def:SAtt} above).

In this paper, we investigate the representational strength of transformers with small weights. Our main result will show a limitation, that without large weights, a transformer cannot take advantage of more than a small number of layers. In other words, we will show that in order to take advantage of the full expressive power of the transformer model, large weights are necessary.

\paragraph{Rank Collapse and Skip Connections}

We will crucially build on the approach of Dong, Cordonnier, and Loukas~\cite{dcl21}, who studied the representational strength of different variants on the transformer architecture through the lens of a notion called \emph{rank collapse}. We say that a model experiences rank collapse if, on any input, the output must always be close to a rank 1 matrix. (See Definition~\ref{def:Res} below for the precise meaning.) Beyond being unable to represent complex concepts, models with rank collapse also have numerous other issues in both training and evaluation \cite{nab+22,rl24,nst24,nunb24,hc24,yx24,bag+25,bgc25}.

The work of \cite{dcl21} highlights \emph{skip connections} (or residual connections) in a transformer network as crucial for avoiding rank collapse. They show that in a Self-Attention Network without skip connections, rank collapse occurs with a doubly exponential rate of convergence. More precisely, if $\beta$ is a bound on the $\ell_1$ norm of the weight matrices of the network, and the network has $L$ layers, then they show the distance to a rank-1 matrix shrinks as \begin{align}O(\beta)^{\frac{3^L - 1}{2}}.\label{eq:oldwork}\end{align} Meanwhile, they observe that networks with skip connections may experience no rank collapse at all. For instance, it is not hard to simulate the \emph{identity} function as a  Self-Attention Network with skip connections (simply set all value weights to $0$, so that only the skip connections are output). In this case, any input which is far from rank-1 will result in an output which is also far from rank-1. They study other mechanisms in transformer networks as well, including multi-layer perceptrons and layer normalization, but find that only skip connections prevents the rank collapse of Equation (\ref{eq:oldwork}).
This result is frequently cited in the literature as evidence of the importance of skip connections \cite{mzw21,nab+22,sabp22,gwdw23,lfz23,klh23,glpr23,kkk+24,jsml25}.

\paragraph{The Importance of Large Weights and Layer Collapse}

We begin with a simple observation: in order for Equation (\ref{eq:oldwork}) to be shrinking as $L$ grows, it is necessary that $\beta$ is small, i.e., that the weights of the network are small. In other words, the result of \cite{dcl21} says that:

{\begin{center} To avoid rank collapse, one needs either skip connections \emph{or} large weights. \end{center}}

In this paper, we prove that Self Attention Networks with skip connections, but with small weights, must suffer from a phenomenon similar to rank collapse which we call \emph{layer collapse}. We say that an $L$-layer Self Attention Network $S$ has layer collapse if there is a nearly equivalent Self Attention Network $S'$ which only has a single layer. In other words, although $S'$ only has one layer, it is still as expressive as $S$, since on any input $X$, the outputs $S(X)$ and $S'(X)$ differ in each entry by at most a small error parameter.

When combined with \cite{dcl21}, our result implies:

{\begin{center} To avoid rank and layer collapse, one needs large weights (skip connections do not suffice). \end{center}}

This challenges the previous popular interpretation of \cite{dcl21}, that skip connections were crucial for the representational strength of the model.

The connection between layer collapse and rank collapse may not be evident from the definitions, but it will become clear in our proofs below. At a high level, we will find that the attention mechanisms in lower layers of the Self Attention Network must exhibit rank collapse (regardless of skip connections), and can thus be removed from the network without substantially changing the output. We will show
\begin{theorem}[Main result, informal] \label{thm:main:informal}
If $S$ is a Self Attention Network whose weight matrices have $\ell_\infty$ norm bounded by $\eta$, then there is a Self Attention Network $S'$ with only one layer, such that on any input $X$, $\| S(X) - S'(X) \|_\infty \leq O(\eta) \cdot \| X \|_\infty$. 
\end{theorem}

In fact, the example from \cite{dcl21} of the identity network with skip connections heavily inspired our definition of layer collapse. That network indeed does not have rank collapse, so we could not hope to prove a version of Theorem~\ref{thm:main} with rank collapse instead of layer collapse. On the other hand, it is essentially not making use of its attention mechanisms; they could be removed without changing the output of the network. Our key idea is to show that, more generally, the attention mechanisms with small weights can be removed from any Self Attention Network, with skip connections, without changing the output of the network by very much.

We make one last comparison: our $\eta$ in Theorem~\ref{thm:main} is a bound on the $\ell_\infty$ norm of the weight matrices, whereas the prior result in Eq.~\eqref{eq:oldwork} above uses parameter $\beta$, which is a bound on the $\ell_1$ norm. Our $\eta$ could thus be quite a bit smaller, and there are thus networks without skip connections where \cite{dcl21} does not imply rank collapse (since $\beta \gg 1$ is too big) but our Theorem~\ref{thm:main} still implies layer collapse (since $\eta \ll 1$ is smaller).

{\bf Roamdap.}
In Section~\ref{sec:related}, we present the related work. In Section~\ref{sec:preli}, we introduce several basic notations and definitions. In Section~\ref{sec:perturbation}, we study perturbation properties of several functions, such as softmax. In Section~\ref{sec:rank_property}, we provide several major rank collapse results. In Section~\ref{sec:conl}, we provide the conclusion of this paper.
\section{Related Work}\label{sec:related}

\paragraph{Low-rank Approximations}

Low rank approximation is a fundamental topic in numerical linear algebra \cite{cw13,nn13,syz23_quantum,syyz23_ellinf}. Many problems require either computationally or analytically finding a low-rank approximation under different settings such as linear and kernel SVMs \cite{gsz25}, tensor regression \cite{swyz21,rsz22,dssw18,djs+19}, low rank approximation with Frobenious norm \cite{cw13,nn13}, weighted low rank approximation~\cite{rsw16,gsyz23,llss25,syyz25}, general norm column subset selection \cite{swz19_neurips_general}, entrywise $\ell_1$ norm low rank approximation \cite{swz17,swz19_neurips_l1}, tensor low rank approximation \cite{swz19}, tensor power method \cite{dsy23}, and matrix CUR decomposition \cite{bw14,swz17,swz19}. Rank collapse and other techniques we use here build on this line of work.

\paragraph{Algorithmic Result for Attention Computations}

The quadratic time complexity of attention mechanisms~\cite{vsp+17} has posed significant computational challenges for long sequences. In response to this problem, a wide range of works have been proposed to reduce computational cost and enhance the scalability of attention mechanisms, including sparsification~\cite{cgrs19,zgd+20,bpc20,hci+21,scl+23,kkf+23,fa23,llss24_sparse,lls+24_conv,hjk+24}, kernel-based approaches~\cite{lz20,ckns20,zhdk23,dswz23,lsss24}, and low-rank methods~\cite{llr16,rsw16,hsw+22_lora,hsk+24,zk24}. 
Additionally, another promising line of research is linear attention~\cite{tbm+19,kvnf20,sis21,dsz23,sdh+23,zfb23,acs+24,llss24_explore,swxl24,zbkr24}, which significantly accelerates traditional softmax attention. Other relevant works have explored important aspects of attention mechanisms, covering topics such as circuit complexity~\cite{cll+24_rope,chl+24_rope,lll+25_hypergraph}, model pruning~\cite{fa23,ssz+24_pruning,slbk24,lls+25_prune}, privacy protection~\cite{lssz24_dp,gsyz24}, regression~\cite{gms23}, half-space reporting (HSR)~\cite{jswz21,cls+24}, and quantum computation~\cite{gsyz23_quantum,zsl24}.

\paragraph{Polynomial Kernels for Attention Acceleration}

With the assumption that model weights are small, polynomial kernels~\cite{as23, as24_neurips} are powerful tools for approximating attention computation in almost linear time complexity, providing promising acceleration for both training and inference of a single attention layer. This approach can be further extended to a wide range of applications. 
For instance, polynomial kernels can provide insights into novel attention mechanisms and model designs, such as modern Hopfield models~\cite{hlsl24}, Diffusion Transformers (DiTs)~\cite{ssz+25,hwsl24}, multi-layer Transformers~\cite{lss+24}, and tensor attention mechanisms~\cite{lssz24_tat,as24_iclr}. These polynomial kernel methods also contribute to efficient and model-utility-preserving fine-tuning of foundation models, such as model adapters~\cite{hsw+22_lora,zhz+23,zjk+23,ghz+23}, multi-task fine-tuning~\cite{gfc21,onr+23,zzj+24}, black-box model tuning~\cite{ssq+22}, and instruction tuning~\cite{ll21,chl+22,mkbh22}. 
Other promising applications include privacy protection in attention computation~\cite{lssz24_dp}, CoT reasoning~\cite{ksl+22,wws+22,sdj+23,zmc+24}, and model calibration~\cite{zwf+21,zlx+23}. Very recently, \cite{ghs+25} further extends the work of \cite{as23} to almost all the regimes of parameter $d$ (see definition of $d$ in Defintion~\ref{def:SAtt}).

\paragraph{Regression Models}

The unprecedented energy consumption in training large-scale ML models has necessitated the development of scalable and efficient ML models~\cite{vrls15,bgms21,mlf+22}. As a simple yet powerful approach to solving various machine learning problems~\cite{bub2015,bpsw21,szz24,ss09}, simple regression models have raised significant concerns in model acceleration, with recent advances from different perspectives, including sketching~\cite{sy21,rsz22,syyz23_ellinf} and pre-conditioning~\cite{ycrm18,kkmr22,syz24}. Our work discusses low-rank approximations in attention mechanisms, while our general insight can be extended to other low-rank method applications, such as accelerated regression models.

\paragraph{Diffusion Models} 

Diffusion models and score-based generative models have achieved remarkable success in generating human-preference-aligned and high-quality visual content~\cite{hja20,ssk+21,bdk+23}. These advances not only benefit vision tasks but also enhance the performance of other applications, such as language modeling~\cite{lgs+23,sas+24}, chemical design~\cite{xpd+23,wxhl24}, and e-commerce~\cite{yww+23,wxf+23,lzw+24}. Relevant works have discussed the theoretical guarantee that diffusion models can be approximated efficiently~\cite{hwsl24,hwl+24,hwl24,gkl+25}. Empirical approaches to accelerate diffusion models have addressed various aspects, such as shortcuts~\cite{fhla24, dnl+24,cgl+25_homo}, parameter pruning~\cite{cskc24,mfw+24}, and lazy computation~\cite{nwz+24,ssz+25}. With these acceleration techniques, diffusion models can be trained on larger-scale data, overcoming inherent limitations such as counting~\cite{hwrl24,cgh+25,ghh+25}, text rendering~\cite{chl+23,txh+24,ghs+25_text}, and adherence to physical constraints~\cite{mcs+25,ghs+25_physical,blx+25}. 
Most diffusion models leverage Transformer backbones for enhanced modelling capability. Our work accelerates attention mechanism computations, significantly benefiting a wide range of diffusion models.

\paragraph{Graph ML Models}

Relational data is prevalent in many real-world scenarios, where graph neural networks (GNNs) are the powerful solutions for mining effective patterns from such relations~\cite{kw17,hyl17,wsz+19}. Recent scalability approaches have widely adopted low-rank approximations, such as sketching~\cite{dra+22,csr+23} and vector quantization~\cite{dkl+21,whz+25}, which can take insights from this paper. These accelerations empower a wide range of applications, including misleading information mitigation~\cite{xwl+22,chl+24_gat}, social network prediction~\cite{fml+19,zgph22}, and human action recognition~\cite{phcc20,lcc+21,flz+21}, while also inspiring advances in multiple aspects of graph learning, such as differential privacy~\cite{llw22,mpp+22}, robustness~\cite{gss+21,djlw22,zwp+22}, and sensitive data removal~\cite{cpm23,zha24,yl25}. A recent work~\cite{zha24} proposes an efficient framework for empowering sensitive data impact removal from trained GNNs with partial retraining, leveraging model utility-aware data partitioning and contrastive sub-model aggregation.

\section{Preliminaries}\label{sec:preli}

In Section~\ref{sec:preli:basic}, we provide basic notation, definitions and facts. In Section~\ref{sec:preli:Res} and Section~\ref{sec:preli:balance}, we define the $\Res$ function and balanced matrix notation which will appear prominently in our constructions. In  Section~\ref{sec:preli:self_attention}, we provide the definition of a multi-layer multi-head Self Attention Network which we study here.

\subsection{Basic Notation and Facts}\label{sec:preli:basic}

For an arbitrary positive integer $n$, we use $[n]$ to represent the set $\{1,2, \cdots, n\}$. We define ${\bf 1}_n$ as a length-$n$ vector where all entries are ones. For any $x \in \R^n$, we use $\exp(x) \in \R^n$ to represent a length-$n$ vector whose $i$-th entry is $\exp(x_i)$. For any vector $x \in \R^n$, we use $x^\top$ to denote its transpose. For a vector $x$, the vector $\ell_2$ norm is denoted by $\| x \|_2$, i.e., $\| x \|_2:=( \sum_{i=1}^n x_i^2 )^{1/2}$. For a vector $x$, we use $\| x \|_{\infty}$ to denote its $\ell_{\infty}$ norm, i.e., $\| x \|_{\infty}:=\max_{i = 1}^n |x_i|$. For a vector $x$, we use $\| x \|_1$ to denote its entrywise $\ell_1$ norm, i.e., $\| x \|_1:= \sum_{i=1}^n |x_i|$. For a matrix, we use $\| A \|_1$ to denote its $\ell_1$ norm, i.e., $\| A \|_1 = \sum_{j,l} |A_{j,l}|$. We use $\| A \|_{\infty}$ to denote its $\ell_{\infty}$ norm, i.e., $\| A \|_{\infty}:= \max_{j,l} |A_{j,l}|$. For a vector $x\in \R^n$, we use $\diag(x)$ to denote a diagonal matrix where $i,i$-th entry on diagonal is $x_i$ for all $i \in [n]$.

\begin{definition}\label{def:alpha_softmax}
For a vector $x \in \R^n$, we define $\alpha(x) := \langle \exp(x) , {\bf 1}_n \rangle$. We define $\soft(x)$ as 
%\begin{align*}
$
    \soft(x) := \alpha(x)^{-1} \exp(x).
$
%\end{align*}
For a matrix $A$, we use the notation $\soft(A)$ to denote that we apply $\soft$ to each row of $A$ individually.
\end{definition}

\begin{fact}[Shift-invariance property of softmax]\label{fac:shift_softmax}
For any vector $x \in \R^n$ and for any fixed scalar $a\in \R$, we have $\soft(x) = \soft(x+a{\bf 1}_n )$.
\end{fact}

\begin{fact}[Norm inequality]\label{fac:norm}
For any matrices $A,B$ we have
\begin{itemize}
    \item $\| A B \|_1 \leq \| A \|_1 \cdot \| B \|_1$.
    \item $\| A B \|_{\infty} \leq \| A \|_{\infty} \cdot \| B \|_{\infty}$.
    \item $\| A B \|_1 \leq \| A \|_1 \cdot \| B \|_{\infty}$.
\end{itemize}
\end{fact}

\subsection{Definitions of \texorpdfstring{$\Res$}{}}\label{sec:preli:Res}

\begin{definition}[$\Res$]\label{def:Res}
Let $Z \in \R^{n \times d}$ denote any matrix, we define function the $\Res: \R^{n \times d} \rightarrow \R^{n \times d}$ as 
%\begin{align*}
$
    \Res(Z):= Z - {\bf 1}_n y^\top
$
%\end{align*}
where $y :=\arg\min_{y \in \R^d} \| Z - {\bf 1}_n y^\top \|_{\infty}$.
\end{definition}

$\Res$ is the key definition behind the notion of rank collapse from prior work~\cite{dcl21}; we will use it here to study layer collapse as well, although we use the $\infty$ norm here in contrast to prior work which uses a $1,\infty$ norm.

\subsection{\texorpdfstring{$\theta$}{}-balance}\label{sec:preli:balance}

We also need a measure of how balanced a matrix is.

\begin{definition}[$\theta$-balance]\label{def:theta_balance}
Given a matrix $E \in \R^{n \times n}$, we define a corresponding matrix $D \in \R^{n \times n}$ to be the diagonal matrix with $D_{i,i} :=\max_{j,l\in [n]} |E_{i,j} - E_{i,l}|$. We say $E$ is $\theta$-balanced, if $\| D \|_{\infty} \leq \theta$. 
\end{definition}

\subsection{Self-Attention Network}\label{sec:preli:self_attention}

\begin{definition}
Let $g$ denote the entry-wise exponentiation function, i.e., for $z\in \R$ we have $g(z) = \exp(z)$, and for a matrix $W$ we have $g(W)_{i, j} = g(W_{i, j})$. Given $A \in \R^{n \times d}$ and weights $Q,K, V \in \R^{d \times d}$, the attention computation can be defined as 
\begin{align*}
    \SAtt_H(X) := \sum_{h=1}^H \underbrace{ D_h^{-1} }_{n \times n} \underbrace{ g( X Q_hK_h^\top X^\top ) }_{n \times n} \underbrace{ X }_{n \times d} \underbrace{ V_h }_{d \times d}
\end{align*}
where $D_h:= \diag( g( X Q_h K_h^\top X^\top ) {\bf 1}_n )$, and where ${\bf 1}_n \in \R^n$ is a length-$n$ vector whose entries are all $1$.
\end{definition}

\begin{definition}
Let $L,H$ denote fixed constants, where $L$ represents the number of layers of the network, and $H$ represents the number of heads per layer. Let $\SAtt_H$ denote the multi-heads version of $\SAtt$ where $H$ is the number of heads. For each $\ell \in [L]$,   $X_{\ell} \in \R^{n \times d}$ denote the $\ell$-th layer input of self-attention network, then we have 
%\begin{align*}
$
    X_{\ell+1} = \SAtt_H(X_{\ell}) + %\mu \cdot
    X_{\ell}.
$
\end{definition}

\section{Perturbation Property}
\label{sec:perturbation}

We now move on to our main proof of layer collapse. We begin by showing that the relevant measure of matrices to not change much when their inputs are perturbed. We will ultimately show that layer collapse occurs because lower layers of the network can be seen as slightly perturbing their inputs.
We study the $\Res$ function in Section~\ref{sec:perturb:Res}, the $\alpha$ function and $\alpha^{-1}$ function in Section~\ref{sec:perturb:alpha}, and the softmax function in Section~\ref{sec:perturb:softmax}.

\subsection{Perturbation Property of Res Function}\label{sec:perturb:Res}

\begin{lemma}\label{lem:matrix_close_then_Res_close}
Let $\Res()$ be defined as Definition~\ref{def:Res}. 
If $\| A - B \|_{\infty} \leq \epsilon$, then 
\begin{align*}
\| \Res(A) - \Res(B) \|_{\infty} \leq  \epsilon.
\end{align*}
\end{lemma}

\begin{proof}
    Let $y \in \R^d$ be the vector such that $\Res(B) = B - {\bf 1}_n y^\top$. Then, 
   $\| \Res(A) - \Res(B) \|_{\infty} \leq \| A - {\bf 1}_N y^\top - \Res(B) \|_{\infty} \leq \| B - {\bf 1}_N y^\top - \Res(B) \|_{\infty} + \| B - A \|_{\infty} \leq 0 + \epsilon$. 
\end{proof}

\subsection{Perturbation Property of Exp Function}\label{sec:perturb:alpha}
\begin{lemma}\label{lem:perturb_alpha}
If the following conditions hold: 
%\begin{itemize}
    %\item 
    Let $a, b \in \R^n$. 
    %\item
    Let $\|b \|_{\infty} \leq \epsilon$. 
%\end{itemize}
Then, we can show
\begin{itemize}
    \item $|\exp(a_i+b_i) - \exp(a_i)| \leq (e^{\epsilon}-1) \cdot \exp(a_i)$.
    \item $|\exp(a_i+b_i) - \exp(a_i)| \leq (e^{\epsilon}-1) \cdot \exp(a_i+b_i)$.
    \item $|\alpha(a+b) - \alpha(a)| \leq (e^{\epsilon}-1) \cdot \alpha(a)$.
    \item $|\alpha(a+b) - \alpha(a)| \leq (e^{\epsilon}-1) \cdot \alpha(a+b)$.
\end{itemize}
\end{lemma}
\begin{proof}
It is easy to see that
\begin{align}\label{eq:exp_minus_1}
    \max\{ |\exp(-b_i) - 1| , | \exp(b_i) - 1 | \} \leq e^{\epsilon}-1
\end{align}

We can show
\begin{align}\label{eq:exp_a_i_b_i_minus_exp_b_i}
    |\exp(a_i+b_i) - \exp(a_i)| = & ~ \exp(a_i) | \exp(b_i) - 1 | \notag \\
    \leq & ~ \exp(a_i) \cdot (e^{\epsilon}-1)
\end{align}
where the first step follows from simple algebra, the second step follows from Eq.~\eqref{eq:exp_minus_1}.

Thus, we have
\begin{align*}
|\alpha(a+b) - \alpha(a)| = & ~ | \langle \exp(a+b) , {\bf 1}_n \rangle - \langle \exp( a ), {\bf 1}_n \rangle | \\
\leq & ~ \sum_{i=1}^n | \exp(a_i+b_i) - \exp(a_i)| \\
\leq & ~ \sum_{i=1}^n \exp(a_i) \cdot (e^{\epsilon}-1) \\
= & ~ (e^{\epsilon}-1) \alpha (a)
\end{align*}
where the second step follows from triangle inequality, the third step follows from Eq.~\eqref{eq:exp_a_i_b_i_minus_exp_b_i}, the last step follows from definition of $\alpha(\cdot)$ function.

Similarly, we can show
\begin{align}\label{eq:exp_a_i_b_i_minus_exp_b_i_2}
|\exp(a_i +b_i) - \exp(a_i) | 
= & ~ \exp(a_i+b_i) | \exp(-b_i) - 1 | \notag \\
\leq & ~ \exp(a_i+b_i) \cdot (e^{\epsilon}-1)
\end{align}
where the first step follows from simple algebra, the second step follows from Eq.~\eqref{eq:exp_minus_1}.

Then, we have
\begin{align*}
    |\alpha(a+b) - \alpha(a)| = & ~ | \langle \exp(a+b) , {\bf 1}_n \rangle - \langle \exp( a ), {\bf 1}_n \rangle | \\
\leq & ~ \sum_{i=1}^n | \exp(a_i+b_i) - \exp(a_i)| \\
\leq & ~ \sum_{i=1}^n \exp(a_i+b_i) \cdot (e^{\epsilon} -1 ) \\
= & ~ (e^{\epsilon}-1) \alpha (a+b)
\end{align*}
where the first step follows from definition of $\alpha$ (Definition~\ref{def:alpha_softmax}), the second step follows from triangle inequality, the third step follows from Eq.~\eqref{eq:exp_a_i_b_i_minus_exp_b_i_2}, and the last step follows from definition of $\alpha$ (Definition~\ref{def:alpha_softmax}).

Thus, we complete the proof.
\end{proof}

\begin{lemma}\label{lem:perturb_alpha_inverse}
If the following conditions hold:
%\begin{itemize}
%    \item
    Let $a, b \in \R^n$. 
    %\item
    Let $\| b \|_{\infty} \leq \epsilon$. % for all $i \in [n]$. 
%\end{itemize}
Then we can show 
\begin{itemize}
    \item $| \alpha(a+b)^{-1} - \alpha(a)^{-1} | \leq (e^{\epsilon}-1) \alpha(a)^{-1}$
    \item $| \alpha(a+b)^{-1} - \alpha(a)^{-1} | \leq (e^{\epsilon}-1) \alpha(a+b)^{-1}$
\end{itemize}
\end{lemma}
\begin{proof}
We can show that
\begin{align*}
    |\alpha(a+b)^{-1} - \alpha(a)^{-1} |
    = & ~ \alpha(a+b)^{-1} \alpha(a)^{-1} | \alpha(a+b) - \alpha(a) | \\
    \leq & ~ \alpha(a+b)^{-1} \alpha(a)^{-1} \cdot (e^{\epsilon}-1) \alpha(a+b) \\
    = & ~ (e^{\epsilon}-1) \alpha(a)^{-1}
\end{align*}
where the first step follows from simple algebra, the second step follows from Lemma~\ref{lem:perturb_alpha}.

Similarly, we can also show $|\alpha(a+b)^{-1} - \alpha(a)^{-1} |\leq (e^{\epsilon}-1) \alpha(a+b)^{-1}$.
\end{proof}

\subsection{Perturbation Property of Softmax Function}\label{sec:perturb:softmax}

\begin{lemma}\label{lem:softmax_a+b_minus_softmax_a_linf}
Let $a, b \in \R^n$. If $|b_i| \leq \epsilon$ for all $i \in [n]$, then, we can show that
\begin{align*}
    \| \soft(a+b) - \soft(a) \|_{\infty} \leq 2(e^{\epsilon}-1)
\end{align*}
\end{lemma}
\begin{proof}

For each $i \in [n]$, we can show
\begin{align*}
& ~ | \alpha(a+b)^{-1} \exp( (a+b)_i ) - \alpha(a)^{-1} \exp( a_i ) | \\
= & ~ | \alpha(a+b)^{-1} \exp( (a+b)_i ) - \alpha(a+b)^{-1} \exp( a_i ) + \alpha(a+b)^{-1} \exp( a_i ) - \alpha(a)^{-1} \exp( a_i ) | \\
\leq & ~ \alpha(a+b)^{-1} | \exp(b_i+a_i) - \exp(a_i) | + \exp(a_i) \cdot |\alpha(a+b)^{-1} - \alpha(a)^{-1}| \\
:= & ~ A_1 + A_2
\end{align*}
where the second step follows from the triangle inequality.

We can upper bound $A_1$ as 
\begin{align*}
A_1 =
& ~ \alpha(a+b)^{-1} \cdot | \exp(a_i+b_i) - \exp(a_i) | \\
\leq & ~ \alpha(a+b)^{-1} \cdot (e^{\epsilon}-1) \exp( a_i+b_i ) \\
\leq & ~ (e^{\epsilon} - 1)
\end{align*}
where the second step follows from Lemma~\ref{lem:perturb_alpha}, the third step follows from $\alpha(x)^{-1} \exp(x_i) \in (0,1)$ for any $x$ and $i$.
 
We can upper bound $A_2$ as
\begin{align*}
A_2 = & ~ \exp(a_i) \cdot | \alpha(a+b)^{-1} - \alpha(a)^{-1} | \\
\leq & ~ \exp(a_i) \cdot (e^{\epsilon}-1) \alpha(a)^{-1} \\
\leq & ~ e^{\epsilon}-1
\end{align*}
where the second step follows from Lemma~\ref{lem:perturb_alpha_inverse}, and the third step follows from $\alpha(x)^{-1} \exp(x_i) \in (0,1)$ for any $x$ and $i$.

Putting everything together, we can show
\begin{align*}
         \| \soft(a+b) - \soft(a) \|_{\infty} 
    = & ~ \max_{i \in [n]} | \alpha(a+b)^{-1} (a+b)_i - \alpha(a)^{-1} a_i | \\
    \leq & ~ 2(e^{\epsilon}-1).
    \end{align*}
Thus, we complete the proof.
\end{proof}

\section{Rank Collapse Property} \label{sec:rank_property}
In Section~\ref{sec:rank:connect}, we present a Lemma which connects $\Res(\SAtt())$ and $\Res()$. In Section~\ref{sec:rank:preserve}, we present our key lemma, a perturbation theorem for a layer of a Transformer. In Section~\ref{sec:rank:main}, we present our main result and proof sketch.

\subsection{The Connection Between \texorpdfstring{$\Res(\SAtt())$}{} and \texorpdfstring{$\Res()$}{}}\label{sec:rank:connect}

We next establish the relationship between $\Res(\SAtt())$ and $\Res()$ in terms of the balance of the inputs.
\begin{lemma}\label{lem:Res_SAtt_X_bound_by_Res_X}
If the following conditions hold:
%\begin{itemize}
    %\item 
    Let $X\in \R^{n \times d}$ denote the input of attention layer.
    %\item 
    Let $\wt{X} = \SAtt(X)$ (see Definition~\ref{def:SAtt} for function $\SAtt$).
    %\item 
    Let $W_q, W_k, W_v \in \R^{d \times d}$ be the weight matrices of $\SAtt$.
    %\item 
    Let $W = W_q W_k^\top$.
    %\item 
    Let $E = \beta \Res(X) W \Res(X)^\top$.
    %\item 
    Suppose that $E$ is a $\theta$-balanced matrix (see Definition~\ref{def:theta_balance}).
    %\item 
    Let $\beta := 1/\sqrt{d_0}$ denote the normalization factor. Let $K: = (e^{\theta} -1) \| W_v \|_{\infty}$.  
Then, we have $\| \Res(\SAtt(X)) \|_{\infty} \leq K \cdot \| \Res(X) \|_{\infty}$.
\end{lemma}

\begin{proof}
The unscaled attention scores are computed as follows
\begin{align*}
     A = \underbrace{ ( X W_q + {\bf 1}_n b_q^\top ) }_{n \times d} \cdot \underbrace{ (X W_k + {\bf 1}_n b_k^\top )^\top }_{d \times n} 
\end{align*}

Recall that $W=W_q W_k^\top$.
For notational convenience, we define
%\begin{align*}
$
 b:=   W_k    b_q  
$.
%\end{align*}

We can use the softmax shift invariance property to remove terms which are constant over the columns and obtain,
%\begin{align*}
$
    A = \underbrace{ X }_{n \times d} \underbrace{ W }_{d \times d} \underbrace{ X^\top }_{d \times n} + \underbrace{ {\bf 1}_n }_{ n \times 1 } \underbrace{ b^\top }_{1 \times d} \underbrace{ X^\top }_{d \times n}
$.
%\end{align*}
 
We define 
$\wt{R}:= \Res( \wt{X} ) \in \R^{n \times d}$ (Recall the definition of the function $\Res()$ in Definition~\ref{def:Res}).

In next equation, we will use the definition of $R$ to simplify $A$. The attention matrix can be written as 
\begin{align}\label{eq:rewrite_A}
    A = & ~ \beta \cdot ( {\bf 1}_n x^\top + R) W ({\bf 1}_n x^\top + R)^\top + \beta \cdot {\bf 1}_n b^\top ( {\bf 1}_n x^\top + R)^\top \notag \\
    = & ~ \beta \cdot ( x^\top W x {\bf 1}_n + R W x + {\bf 1}_n b^\top x   ) {\bf 1}_n^\top + \beta \cdot (R W R^\top + {\bf 1}_n x^\top W R^\top +  {\bf 1}_n b^\top R^\top)
\end{align}

Using Fact~\ref{fac:shift_softmax}, we can remove the first term in the above equation since it is constant across columns. We thus have that the following equation for $P = \soft(A) \in \R^{n \times n}$
\begin{align}\label{eq:rewrite_P}
    P = & ~ \soft ( \beta R W R^\top + {\bf 1}_n r^\top  ) \notag \\
    = & ~ \soft( E + {\bf 1}_n r^\top )
\end{align}
where the first step follows from $r = \beta R( W^\top x + b) \in \R^n$, the second step follows from setting $E= \beta R W R^\top \in \R^{n \times n}$.

To continue the proof, we also set $\wt{A} = {\bf 1}_n r^\top \in \R^{n \times n}$, the input reweighted by the attention probabilities $PX$ will be entry-wisely upper bounded as follows
\begin{align}\label{eq:rewrite_PX}
    PX = & ~  P( {\bf 1}_n x^\top + R ) \notag \\
    = & ~ {\bf 1}_n x^\top + P R \notag \\
    = & ~ {\bf 1}_n x^\top + \soft ( {\bf 1}_n r^\top + E ) R \notag \\
    \leq & ~ {\bf 1}_n x^\top + (I + e^{D} -I ) {\bf 1}_n \soft(r)^\top R \notag \\
    = & ~ {\bf 1}_n (x^\top \soft(r)^\top R) + (e^{D} - I) {\bf 1}_n \soft(r)^\top R
\end{align}
where the first step follows from definition of $R$, the second step follows from $P {\bf 1}_n = {\bf 1}_n$, the third step follows from Eq.~\eqref{eq:rewrite_P}, the forth step follows from Lemma~\ref{lem:lemma_A.3} and $e^D$ is diagonal matrix where the $i,i$-th entry on diagonal is $e^{D_{i,i}}$.

Therefore, the entry-wise distance of the output of the self-attention layer $\SAtt(X) = P X W_v$ from being constant across token is at most:
\begin{align*}
| [ \SAtt(X) - {\bf 1}_n \wt{r}^\top ]_{i,j} |
= & ~ | [ P X W_v - {\bf 1}_n \wt{r}^\top ]_{i,j} | \\
\leq & ~ (e^{\theta}-1) \cdot | [ D {\bf 1}_n \soft(r)^\top R W_v ]_{i,j} |
\end{align*}
where the second step follows from $\wt{r}= (x + R^\top \soft(r) ) W_v$ and Eq.~\eqref{eq:rewrite_PX}.

Now we bound the right hand side of the above inequality.

For $\| \cdot \|_{\infty}$, we can show
\begin{align}\label{eq:R'_infty}
\| \SAtt(X) - {\bf 1}_n (r')^\top \|_{\infty} 
\leq & ~ (e^\theta- 1) \| D {\bf 1}_n \|_{\infty} \cdot \| \soft(r)^\top R W_v \|_{\infty} \notag \\
\leq & ~ (e^\theta- 1) \| D {\bf 1}_n \|_{\infty} \| R \|_{\infty} \| W_v \|_{\infty} \notag \\
\leq & ~ (e^\theta- 1) \| R \|_{\infty} \cdot \| W_v \|_{\infty},
\end{align}
where the last step follows from Definition~\ref{def:theta_balance}.

Note that $R' = \Res(\SAtt(X))$ and $R = \Res(X)$ and using the definition of $K$ in Lemma statement, we can show
\begin{align*}
    \| \Res( \SAtt( X ) ) \|_{\infty} \leq K \cdot \| \Res(X) \|_{\infty}
\end{align*}
Thus, we complete the proof.
\end{proof}

\subsection{Perturbation of One Transformer Layer}\label{sec:rank:preserve}

\begin{lemma}[Single Head]\label{lem:rank1_single_head}
    %\begin{itemize}
        %\item 
        Let $X \in \R^{n\times d}$.
        %\item
        Let $A = \soft_1(X)$ (Recall that $\soft()$ function is defined as Definition~\ref{def:alpha_softmax}. Note that $\soft_1$ and $\soft_2$ are two different instantiations with different $W_k,W_q,W_v$ weights).
        %\item 
        Let $B = X + A$. 
        %\item 
        Suppose $\| \Res(A)\|_{\infty} \leq 
        K \cdot \|\Res(X)\|_{\infty} \leq \epsilon$. (We remark that this condition will hold due to Lemma~\ref{lem:Res_SAtt_X_bound_by_Res_X}; here $K$ is as defined in Lemma~\ref{lem:Res_SAtt_X_bound_by_Res_X}). 
        %\item 
        Let $g(\epsilon) := 2(e^{\epsilon}-1)$ and let $\epsilon_0 = 2 g(2\epsilon)$.
    %end{itemize}
    Then we can show
    %\begin{itemize}
      \begin{align*}
        \|\soft_2(B) - \soft_2(X)\|_{\infty} \leq \epsilon_0.
        \end{align*}
    %\end{itemize}
\end{lemma}

\begin{proof}
    Let $R_X = \Res(X)$ so that $X = R_X + y_X {\bf 1}_d^\top$ for some vector $y_X \in \R^n$. Using Fact~\ref{fac:shift_softmax}, we can show that 
    \begin{align}\label{eq:soft_X_is_soft_R_X}
    \soft_1(X) = \soft_1(R_X).
    \end{align}

    Let $R_A = \Res(A)$ so that $A = R_A + y_A {\bf 1}_d^\top$ for some vector $y_A \in \R^n$. Using Fact~\ref{fac:shift_softmax}, we can show that $\soft(A) = \soft(R_A)$.

    Let $R_B = \Res(B)$ so that $B= R_B + y_B {\bf 1}_d^\top$ for some vector $y_B \in \R^n$. Using Fact~\ref{fac:shift_softmax}, we can show that
    %\begin{align*}
    $
        \soft_2(B) = \soft_2(R_B).
    $
    %\end{align*}

    Next we will show that $\| \soft(R_X+R_A) - \soft(R_X) \|_{\infty}$ is small.

    Let us consider the vector version $\| \soft(a+b) - \soft(a) \|_{\infty}$. Note that if $ |b_i| \leq \epsilon $, then 
 using Lemma~\ref{lem:softmax_a+b_minus_softmax_a_linf}, we can show
    \begin{align*}
     \| \soft(a+b) - \soft(a) \|_{\infty}
    \leq g(\epsilon)
    \end{align*}

Thus, as long as $\| R_A \|_{\infty} \leq \epsilon$, then using Lemma~\ref{lem:softmax_a+b_minus_softmax_a_linf}, we have
\begin{align}\label{eq:soft_R_X_R_A_minus_soft_R_X}
    \| \soft_2(R_X + R_A ) - \soft_2(R_X) \|_{\infty} \leq g(\epsilon)
\end{align}

We can show $R_X = \Res(X) = \Res(B - A) = \Res(B - R_A)$. Then, we know $ \| R_X - R_B \|_{\infty} \leq \| \Res(B-R_A) - \Res(B) \|_{\infty} \leq \| R_A \|_{\infty} \leq \epsilon$.

Recall $B = X + A$, and $\| R_{A} \|_{\infty}\leq \epsilon$ then we know
\begin{align*}
 \| R_X + R_A - R_B \|_{\infty} \leq \| R_X- R_B \|_{\infty} + \| R_A \|_{\infty} \leq 2 \| R_A \|_{\infty} \leq  2 \epsilon.
\end{align*}

Since $\| R_X + R_A - R_B \|_{\infty} \leq 2\epsilon$,
then using Lemma~\ref{lem:softmax_a+b_minus_softmax_a_linf}, we have
\begin{align}\label{eq:soft_R_X_R_A_minus_soft_R_B}
    \| \soft_2(R_X+R_A) - \soft_2(R_B) \|_{\infty} \leq g(2\epsilon)
\end{align}

Then, we can show
\begin{align*}%\label{eq:bound_softmax_B_and_X}
    & ~ \| \soft_2(B) - \soft_2(X) \|_{\infty} \notag \\
    = & ~ \| \soft_2(B) - \soft_2(R_X) \|_{\infty} \notag  \\
    = & ~ \| \soft_2(R_B) - \soft_2(R_X) \|_{\infty} \notag \\
    \leq & ~ \| \soft_2(R_B) - \soft_2(R_X+R_A) \|_{\infty} + \| \soft_2(R_X+R_A) - \soft_2(R_X) \|_{\infty}  \notag  \\
    \leq & ~ g(2\epsilon) + g(\epsilon) \notag\\
    \leq & ~ 2g(2\epsilon)
\end{align*}
where the first step follows from $\soft(X)= \soft(R_X)$, the second step follows from $\soft(R_B) =\soft(B)$, the third step follows from triangle inequality, the forth step follows from Eq.~\eqref{eq:soft_R_X_R_A_minus_soft_R_X} and Eq.~\eqref{eq:soft_R_X_R_A_minus_soft_R_B}, and the last step follows from $g$ is monotone.\qedhere

\end{proof}

\subsection{Putting it All Together}\label{sec:rank:main}
\begin{theorem}[Restatement of Theorem~\ref{thm:main:informal}]\label{thm:main}
    Suppose $S$ is a $\SAtt$ with residuals, with the property that for every attention head in every one of its layers, the weight matrices $W_q, W_k, W_v \in \R^{d \times d}$ all have the bound $\| W_q \|_{\infty}, \| W_k \|_{\infty} , \| W_v \|_{\infty} \leq \eta $. 
    Then, there exists a $\SAtt$ with residuals $S'$ with just one layer so that, for any $X \in \R^{n \times d}$, we have $\| S(X) - S'(X) \|_{\infty} \leq  O(\eta)  \cdot \| X \|_{\infty}$.
\end{theorem}

\begin{proof}[Proof Sketch]

We'll show what to do to delete one layer, then repeat that $L - 1$ times to get down to one layer. When we delete the first layer, Lemma~\ref{lem:rank1_multi_head} (which is the version of Lemma~\ref{lem:rank1_single_head} which deals with multiple heads)  says that the output of the second layer will differ by at most $O(\eta \cdot  \epsilon_0)$, where $\epsilon_0 = O(1) \cdot \| X \|_{\infty}$ is the constant from Lemma~
\ref{lem:Res_SAtt_X_bound_by_Res_X} and Lemma~\ref{lem:rank1_single_head}, and $X$ is the input of first layer of network. Therefore, by applying Lemma~\ref{lem:softmax_a+b_minus_softmax_a_linf} iteratively to each layer, it follows that the outputs of all subsequent layers will also change by at most $O(\eta \cdot  \epsilon_0)$. In particular, the final output will differ by at most $O(\eta \cdot  \epsilon_0)$. We finally repeat this $L - 1$ times to remove all but one layer and get the final error. We defer further proof details to the Appendix due to space limitations. 
\end{proof}

\section{Conclusion}\label{sec:conl}

We have shown that Self Attention Networks must experience layer collapse unless they have large attention weights, even if they have skip connections. This shows that large weights, and not residual or skip connections, are the key to the expressive power of Self Attention Networks. 
Moving forward, these insights suggest that efforts to improve model capability should emphasize principled parameter scaling and regularization strategies, rather than relying on more elaborate residual topologies.

\ifdefined\isarxiv

\else
\bibliographystyle{alpha}
\bibliography{ref}
\fi

%%% The part below is the NeurIPS checklist
\newpage
\ifdefined\isarxiv
\else
\input{checklist}
\fi

%%% The part below is the appendix

\newpage
\onecolumn
\appendix

\begin{center}
    \textbf{\LARGE Appendix }
\end{center}

%%% This file is the structure for appendix content
%%% TeX files for body contents should be named as:
%%% 50_xxxx.tex
%%% 51_xxxx.tex
%%% ...

{\bf Roadmap.} In Section~\ref{sec:app_preli}, we provide several simple definitions. In Section~\ref{sec:perturb_E}, we show more perturbation properties for the softmax matrix. In Section~\ref{sec:multiple_heads}, we provide the proofs related to network layers with multiple attention heads. In Section~\ref{sec:multiple_layers}, we prove our main Theorem. Finally, in Section~\ref{sec:limitation}, we discuss the limitation of this work, and in Section~\ref{sec:impact}, we provide the broader impact of this work.

\section{Preliminaries}\label{sec:app_preli}

Note that $\soft(X)$ function defined above is ignoring the effect of the weights $W_v$. Here we incorporate them in another function $\softmv(X)$ (which is also usually called self-attention).
\begin{definition}\label{def:softmv}
Let $W_q, W_k$ be weights being used in $\soft$. Let $W_v$ denote the extra weights that will be used in $\softmv$.
We define $\softmv(X)$ as follows
\begin{align*}
    \softmv(X) := \soft(X) X W_v.
\end{align*}
\end{definition}
Next we define a very useful parameter $\epsilon_{\ell}$ which captures the Lipschitz and layer norm property of every layer.
\begin{definition}\label{def:epsilon_ell}
Let all layers' weights are bounded, i.e, $\| W_q\|_{\infty}, \| W_k \|_{\infty}, \| W_v \|_{\infty} \leq \eta$. Let $X_0$ denote the first layer input of entire neural network and it is bounded $\| X_0 \|_{\infty} \leq \phi_0$. Let $H$ denote the number of heads. For each layer $\ell \in [L]$, we define a parameter $\epsilon_{\ell}:=2\eta \phi_0 (1+H \eta)^{\ell} $.
\end{definition}

\section{Perturbation Property of Softmax Matrix }\label{sec:perturb_E}

\begin{lemma}[]\label{lem:lemma_A.3}
If the following conditions hold
\begin{itemize}
    \item Let $P = \soft(A)$ (see Definition~\ref{def:alpha_softmax} for function $\soft()$).
    \item Let $\wt{A} = A -E$.
    \item Let $\wt{P} = \soft( \wt{A} )$. 
    \item Let $D$ be defined as Definition~\ref{def:theta_balance}, i.e., $D_{i,i}:=\max_{j,l \in [n]} | E_{i,j} - E_{i,l} |$
\end{itemize}
Then we can show, for all $i \in [n], j \in [n]$
\begin{align*}
e^{-D_{i,i}} \wt{P}_{i,j} \leq P_{i,j} \leq e^{D_{i,i}} \wt{P}_{i,j}.
\end{align*}
 
\end{lemma}
\begin{proof}
Let us start by the definition of $P$, for each $i \in [n]$, $j \in [n]$
\begin{align*}
    P_{i,j} 
    = & ~ (\soft(A))_{i,j} \\
    = & ~ (\soft(\wt{A} + E) )_{i,j}\\
    = & ~ \frac{ \exp( \wt{A}_{i,j} + E_{i,j} ) }{ \sum_{l=1}^n \exp( \wt{A}_{i,l} + E_{i,l} ) } \\
    = & ~ \frac{ \exp( \wt{A}_{i,j} )  )  }{ \sum_{l=1}^n \exp( \wt{A}_{i,l} ) \exp( E_{i,l} - E_{i,j} ) }
\end{align*}
where the first step follows from definition of $P$, the second step follows from definition of $A$, the third step follows from definition of softmax (Definition~\ref{def:alpha_softmax}), and the last step follows from property of $\exp$.

We define $D_{i,i}:=\max_{j,l \in [n]} | E_{i,j} - E_{i,l} |$. 
We have that
\begin{align*}
 P_{i,j} \in [ \wt{P}_{i,j}  \exp( - D_{i,i}  ) , \wt{P}_{i,j}  \exp( D_{i,i} ) ]
\end{align*}
Thus, we complete the proof.
\end{proof} 

\begin{lemma}\label{lem:theta_is_1}
If the following conditions hold
\begin{itemize}
    \item Let $W_q, W_k, W_v$ be the matrix that $\| W_q \|_{\infty}, \| W_k \|_{\infty}, \| W_v \|_{\infty} \leq \eta$.
    \item Let $W = W_q W_k^\top$.
    \item Let $E = \beta \Res(X) W \Res(X)^\top$.
    \item Let $\beta$ satisfy that $\beta \leq 1/( \| \Res(X) \|_{\infty}^2 \eta^2 )$.
\end{itemize}
Then, we have $E$ is $\theta$-balanced with $\theta=1$.
\end{lemma}
\begin{proof}
First, note that $\| W \|_{\infty} \leq \| W_q \|_{\infty} \cdot \| W_k \|_{\infty} \leq \eta^2$.

We can show
\begin{align*}
    \| E \|_{\infty} 
    \leq & ~ \beta \cdot \| \Res(X) \|_{\infty}^2 \cdot \| W \|_{\infty}  \\
    \leq & ~ \beta \cdot \| \Res(X) \|_{\infty}^2 \cdot \eta^2 \\
    \leq & ~ 1
\end{align*}
Thus, we complete the proof.
\end{proof}

\section{Multiple Heads}\label{sec:multiple_heads}

Here, we generalize the proof of Lemma~\ref{lem:rank1_single_head} to multiple heads. Note that Lemma~\ref{lem:rank1_single_head} presented a simplified proof by ignoring the effects of $X W_v$, and thus automatically assuming $n=d$. In this section we remove that condition and prove the result for general $n$ and $d$. In Section~\ref{sec:multiple_heads:main}, our goal is to prove Lemma~\ref{lem:rank1_multi_head}, which is the multiple heads version of Lemma~\ref{lem:rank1_single_head}. In Section~\ref{sec:multiple_heads:conditions}, we show that several required conditions in Lemma~\ref{lem:rank1_multi_head} are satisfied.

\subsection{Multiple Heads For Skipping One Layer}\label{sec:multiple_heads:main}
In the next Lemma~\ref{lem:rank1_multi_head}, we will put the effect of $\softmv$ back. We remark that the major idea of the proof remains the same as Lemma~\ref{lem:rank1_single_head}.
\begin{lemma}[Multiple Heads version of Lemma~\ref{lem:rank1_single_head}]\label{lem:rank1_multi_head}
If the following conditions hold,
    \begin{itemize}
        \item Let $H$ denote the number of heads.
        \item Note that $\soft_1$ and $\soft_2$ are two different instantiations with different $W_k, W_q, W_v$ weights.
        \item Let $X \in \R^{n\times d}$.
        \item $A_i = \softmv_{1,i}(X) \in \R^{n \times d}$ for $i \in [H]$. (Let $\softmv$ be defined as Definition~\ref{def:softmv})
        \item $B = X + \sum_{i=1}^H A_i \in \R^{n \times d}$.
        \item $\| \Res(A_i)\|_{\infty} \leq K \cdot \| \Res(X) \|_{\infty} \leq \epsilon $ for all $i \in [H]$. (We remark that this condition will hold due to Lemma~\ref{lem:Res_SAtt_X_bound_by_Res_X}; here $K$ is as defined in Lemma~\ref{lem:Res_SAtt_X_bound_by_Res_X})
        \item Let $g(\epsilon) := 2(e^{\epsilon} - 1)$.
        \item Let $W_v$ satisfy that $\| (B-X) W_v \|_{\infty} \leq H \cdot \epsilon$ and $\| X W_v \|_{\infty} \leq 1$
        \item Let $\epsilon_0 = 3g(2H\epsilon)$ 
    \end{itemize}
    Then we can show
    \begin{itemize}
        \item {\bf Part 1.} $\|\soft_2(B) - \soft_2(X)\|_{\infty} \leq   \epsilon_0$
        \item {\bf Part 2.} $\| \softmv_2(B) - \softmv_2(X) \|_{\infty} \leq \epsilon_0$
    \end{itemize}
\end{lemma}
\begin{proof}

{\bf Proof of Part 1.}

    Let $R_X = \Res(X)$ so that $X = R_X + y_X {\bf 1}_d^\top$ for some vector $y_X \in \R^n$. 
    
    Using Fact~\ref{fac:shift_softmax}, we can show that 
    \begin{align}\label{eq:soft_X_is_soft_R_X:H}
    \soft_1(X) = \soft_1(R_X).
    \end{align}

To notataionaly help in our proof, we define the prefix sums of matrices $A_0, A_1, \cdots , A_i \in \R^{n \times d}$ as
\begin{align*}
    A_{[i]}:=\sum_{j=0}^i A_i
\end{align*}
where $A_0$ is an artificial matrix that has $0$ everywhere.

   For each $i \in [H]$, let $R_{A_{[i]}} = \Res(A_{[i]})$ so that $A_{[i]} = R_{A_{[i]}} + y_{A_{[i]}} {\bf 1}_d^\top$ for some vector $y_{A,{[i]}} \in \R^n$. 
   
   Using Fact~\ref{fac:shift_softmax}, we can show that 
   \begin{align*}
   \soft(A_{[i]}) = \soft(R_{A_{[i]}}).
   \end{align*}

    Let $R_B = \Res(B)$ so that $B= R_B + y_B {\bf 1}_d^\top$ for some vector $y_B \in \R^n$. Using Fact~\ref{fac:shift_softmax}, we can show that
    \begin{align}\label{eq:soft_B_is_soft_R_B:H}
        \soft_2(B) = \soft_2(R_B).
    \end{align}

    Let us consider the vector version $\| \soft(a+b) - \soft(a) \|_{\infty}$. Note that if $ \|b\|_{\infty} \leq \epsilon $, then 
 using Lemma~\ref{lem:softmax_a+b_minus_softmax_a_linf}, we can show
    \begin{align*}
     \| \soft(a+b) - \soft(a) \|_{\infty}
    \leq g(\epsilon)
    \end{align*}
 
Since $\| R_{A_{[i]}} - R_{ A_{[i-1]} } \|_{\infty} \leq \epsilon$ for all $i \in [H]$, then using Lemma~\ref{lem:softmax_a+b_minus_softmax_a_linf}, we have: for each $i \in [H]$
\begin{align}\label{eq:soft_R_X_R_A_minus_soft_R_X:H}
    \| \soft_2(R_X + R_{A_{[i]}} ) - \soft_2(R_X +R_{A_{[i-1]}} ) \|_{\infty} \leq g(\epsilon)
\end{align}

We can show $R_X = \Res(X) = \Res(B - A_{[H]}) = \Res(B - R_{A_{[H]}})$. 

Then, we know 
\begin{align}\label{eq:R_X_minus_R_B_at_most_R_A}
\| R_X - R_B \|_{\infty} 
= & ~ \| \Res(B-R_{A_{[H]}}) - \Res(B) \|_{\infty} \notag\\
\leq & ~ \| R_{A_{[H]}} \|_{\infty} 
\end{align}
where the first step follows from $R_X = \Res(B-R_{A_{[H]}})$ and $R_B = \Res(B)$.

Recall $B = X + A$, and $\| R_{A} \|_{\infty}\leq \epsilon$ then we know
\begin{align*}
 \| R_X + R_A - R_B \|_{\infty} 
 \leq & ~ \| R_X- R_B \|_{\infty} + \| R_{A_{[H]}} \|_{\infty} \\
 \leq & ~ 2 \| R_{A_{[H]}} \|_{\infty} \\
 \leq & ~ 2 H \epsilon,
\end{align*}
where the first step follows from triangle inequality, the second step follows from $\| R_X - R_B \|_{\infty} \leq \| R_{A_{[H]}} \|_{\infty}$, and the last step follows from $\| R_{A_{[H]}} \|_{\infty} \leq H \epsilon$.

Since $\| R_X + R_{A,H} - R_B \|_{\infty} \leq 2H\epsilon$,
then using Lemma~\ref{lem:softmax_a+b_minus_softmax_a_linf}, we have
\begin{align}\label{eq:soft_R_X_R_A_minus_soft_R_B:H}
    \| \soft_2(R_X+R_{A,H}) - \soft_2(R_B) \|_{\infty} \leq g(2H\epsilon)
\end{align}

Then, we can show
\begin{align}\label{eq:soft_2_B_minus_X} 
    & ~ \| \soft_2(B) - \soft_2(X) \|_{\infty} \notag \\
    = & ~ \| \soft_2(B) - \soft_2(R_X) \|_{\infty} \notag  \\
    = & ~ \| \soft_2(R_B) - \soft_2(R_X) \|_{\infty} \notag \\
    \leq & ~ \| \soft_2(R_B) - \soft_2(R_X+R_{A,H}) \|_{\infty}\notag  \\
    & ~ + \sum_{i=1}^{H-1} \| \soft_2( R_X + R_{A,i} ) - \soft_2( R_X + R_{A,i-1} ) \|_{\infty} \notag \\
    \leq & ~ g(2H\epsilon) + H \cdot g(\epsilon) \\
    \leq & ~ 2g(2H\epsilon)\notag 
\end{align}
where the first step follows from $\soft(X)= \soft(R_X)$ (see Eq.~\eqref{eq:soft_X_is_soft_R_X:H}), the second step follows from $\soft(R_B) =\soft(B)$ (see Eq.~\eqref{eq:soft_B_is_soft_R_B:H}), the third step follows from triangle inequality, the forth step follows from Eq.~\eqref{eq:soft_R_X_R_A_minus_soft_R_X:H} and Eq.~\eqref{eq:soft_R_X_R_A_minus_soft_R_B:H}, and the last step follows from property of function $g$. 

{\bf Proof of Part 2.}

We can show that
\begin{align*}
& ~ \| \softmv_2(B) - \softmv_2(X) \|_{\infty}  \\
= & ~ \| \soft_2(B) B W_v - \soft_2(X) X W_v \|_{\infty}  \\
\leq & ~ \| \soft_2(B) B W_v - \soft_2(B) X W_v \|_{\infty} + \| \soft_2(B) X W_v - \soft_2(X) X W_v \|_{\infty} \\
\leq & ~ \| \soft_2(B) \|_{\infty} \cdot \| (B - X ) W_v \|_{\infty} + \| \soft_2(B) - \soft_2(X)  \|_{\infty} \cdot \| X W_v \|_{\infty} \\
\leq & ~ H \epsilon +g(2H \epsilon) + H \cdot g(\epsilon) \\
\leq & ~ 2 g(2H \epsilon)
\end{align*}
 where the second step follows from triangle inequality, the third step follows from Fact~\ref{fac:norm}, and the forth step follow from Eq.~\eqref{eq:soft_2_B_minus_X} , where the last step follows from $\epsilon \leq g(\epsilon)$ and $2H g(\epsilon) \leq g(2H \epsilon)$.
\end{proof}

\subsection{Conditions in Lemma~\ref{lem:rank1_multi_head} are Satisfied}\label{sec:multiple_heads:conditions}
Here we will show that the three conditions in Lemma~\ref{lem:rank1_multi_head} will be satisfied for each layer $\ell$.
\begin{itemize}
    \item $\| \Res(A_i) \|_{\infty} \leq K \cdot \| \Res(X) \|_{\infty} \leq \epsilon$ (where $K:= (e^{\theta}-1) \| W_v \|_{\infty}$, definition of $K$ recall Lemma~\ref{lem:Res_SAtt_X_bound_by_Res_X}). Here $\theta =1$ due to Lemma~\ref{lem:theta_is_1}
    \item $\| (B-X) W_v \|_{\infty} \leq H \cdot g(\epsilon)$
    \item $\| X W_v \|_{\infty} \leq 1$
\end{itemize} 

\begin{lemma}
If the following conditions hold
\begin{itemize}
    \item $\epsilon_{\ell}:=2\eta \phi_0 (1+H \eta)^{\ell} $. (see Definition~\ref{def:epsilon_ell})
    \item Let $\eta \in (0,1]$.
    \item Let $\epsilon_{\ell} \in (0, 1)$.
\end{itemize}
Then, we can show
\begin{itemize}
    \item Part 1. $K \cdot \| \Res(X_{\ell}) \|_{\infty} \leq \epsilon_{\ell}$
    \item Part 2. $\| (B- X_{\ell}) W_v \|_{\infty} \leq H \epsilon_{\ell}$
    \item Part 3. $\| X_{\ell} W_v \|_{\infty} \leq 1$
\end{itemize}
\end{lemma}
\begin{proof}
{\bf Proof of Part 1.}

We can show that
\begin{align*}
K \cdot \| \Res(X_{\ell}) \|_{\infty} 
\leq & ~ 2 \eta \| \Res(X_{\ell}) \|_{\infty} \\
\leq & ~ 2 \eta \| X_{\ell} \|_{\infty} \\
\leq & ~ 2 \eta \phi_0 \cdot (1+ H \eta)^{\ell} \\
= & ~ \epsilon_{\ell}
\end{align*}
where the first step follows from $\theta=1$ and $\| W_v\|_{\infty} \leq \eta$, the third step follows from Lemma~\ref{lem:layer_norm}, and the last step follows from definition $\epsilon_{\ell}$

{\bf Proof of Part 2.}
\begin{align*}
    \| (B- X_{\ell}) W_v \|_{\infty}
    \leq & ~ \eta \cdot \| B - X_{\ell} \|_{\infty} \\
    = & ~ \eta \cdot \sum_{i=1}^H \| \softmv_i(X_{\ell}) \|_{\infty} \\
    \leq & ~ \eta \cdot \sum_{i=1}^H \| \soft_i(X_{\ell}) X_{\ell} W_{v,i} \|_{\infty} \\
    \leq & ~ \eta \cdot \sum_{i=1}^H \| \soft_i(X_{\ell}) \|_{\infty} \cdot \| X_{\ell} \|_{\infty} \cdot \| W_{v,i} \|_{\infty} \\
    \leq & ~ \eta^2 H \cdot \| X_{\ell} \|_{\infty} \\
    \leq & ~ 0.5 \eta H \cdot \epsilon_{\ell} \\
    \leq & ~ H \epsilon_{\ell}
\end{align*} 
where the first step follows from $\| W_v \|_{\infty} \eta$, the second step follows from definition of $B$, the third step follows from definition of $\softmv$, the forth step follows from Fact~\ref{fac:norm}, the fifth step follows from $\| \soft() \|_{\infty} \leq 1$ and $\| W_{v,i} \|_{\infty} \eta$, and the last step follows from $\eta \leq 1$.

{\bf Proof of Part 3.}

We can show
\begin{align*}
    \| X_{\ell} W_v \|_{\infty}
    \leq & ~ \eta \| X_{\ell} \|_{\infty} \\
    \leq & ~ \eta \phi_0 (1+H \eta)^{\ell} \\
    \leq & ~ \epsilon_{\ell} \\
    \leq & ~ 1
\end{align*}
where the second step follows from Lemma~\ref{lem:layer_norm}, the third step follows from choice of $\epsilon_{\ell}$ last step follows from Lemma statement condition.

\end{proof}

\section{Multiple Layers} \label{sec:multiple_layers}%{sec:proof_main}

In Section~\ref{sec:multiple_layers:main}, we provide the proof of our main theorem. In Section~\ref{sec:multiple_layers:lipschitz}, we provide the Lipshitz property of several key functions being used in our proofs. In Section~\ref{sec:multiple_layers:layer_lipschitz}, we prove the Lipschitz property for each layer of our Self Attention Network. Finally, in Section~\ref{sec:multiple_layers:layer_norm}, prove the norm of each layer in the Self Attention Network is not increasing much.

\subsection{Proof of Theorem~\ref{thm:main}}\label{sec:multiple_layers:main}

\begin{proof}

We define
\begin{align*}
{X}_{\ell}^{\ell_0} = & {B}_{\ell}^{\ell_0} 
\end{align*}
Then, we define
\begin{align*}
 {B}_{\ell}^{\ell_0} =
 \begin{cases}
{X}_{\ell-1}^{\ell_0} + \sum_{i=1}^H A_{\ell-1,i}^{\ell_0}, & \mathrm{if} \ell  \leq \ell_0; \\
 \sum_{i=1}^H A_{\ell-1,i}^{\ell_0} & \mathrm{otherwise}.
 \end{cases}
\end{align*}
Let $\softmv()$ function be defined as Defintion~\ref{def:softmv}. We define
\begin{align*}
 A_{\ell-1,i}^{\ell_0} = & ~ \softmv_{\ell-1,i}^{\ell_0}(X_{\ell-1}^{\ell_0})
\end{align*}

Note that the notation $B^{0}_L$ means we have residual in every layer, whereas the notation $B^{\ell_0}_L$ means we don't have a residual connection from layer $\ell_0$ to layer $L$.

Let $\epsilon_{\ell}$ be defined as Definition~\ref{def:epsilon_ell}. 
Let $\delta:=\max_{\ell \in [L]} 2g(2H\epsilon_{\ell})$.
Using Lemma~\ref{lem:rank1_multi_head}, we can show 
for all $\ell \in [L]$, 
\begin{align*}
\| \softmv_{\ell}(B^{\ell-1}_{\ell}) - \softmv_{\ell}(B^{\ell}_{\ell}) \|_{\infty} \leq \delta
\end{align*}

Let $C:=\max_{\ell \in [L]} 3 \eta (\epsilon_{\ell}^2 + 1)$.

Then we can show
\begin{align*}
   & ~ \| \softmv_2(B^0_{2}) - \softmv_2(B^{2}_{2}) \|_{\infty} \\
    \leq & ~ \| \softmv_2(B^0_{2}) - \softmv_2(B^{1}_{2}) \|_{\infty} + \| \softmv_2(B^1_{2}) - \softmv_2(B^{2}_{2}) \|_{\infty} \\
    \leq & ~ C \cdot \| \softmv_1(B^0_{1}) - \softmv_1(B^{1}_{1}) \|_{\infty} + \| \softmv_2(B^1_{2}) - \softmv_2(B^{2}_{2}) \|_{\infty} \\
    \leq & (C+1) \delta
\end{align*}
where the first step follows from triangle inequality, the second step follows from the fact that one layer of the network is $C$-Lipschitz (see Lemma~\ref{lem:softmv_lipschitz}), and the last step follows from merging the errors.

For three layers, we have
\begin{align*}
     & ~ \| \softmv_3(B^0_{3}) - \softmv_3(B^{3}_{3}) \|_{\infty} \\
     \leq & ~  \| \softmv_3(B^0_{3}) - \softmv_3(B^{1}_{3}) \|_{\infty} \\
     & ~ +  \| \softmv_3(B^1_{3}) - \softmv_3(B^{2}_{3}) \|_{\infty} \\
     & ~ +  \| \softmv_3(B^2_{3}) - \softmv_3(B^{3}_{3}) \|_{\infty} \\
     \leq & ~ C^2 \cdot \| \softmv_1(B^0_{1}) - \softmv_1(B^{1}_{1}) \|_{\infty}\\
     & ~ + C \cdot \| \softmv_2(B^1_{2}) - \softmv_2(B^{2}_{2}) \|_{\infty} \\
     & ~ +  \| \softmv_3(B^2_{3}) - \softmv_3(B^{3}_{3}) \|_{\infty} \\
     \leq & ~ C^2 \delta + C \delta + \delta \\
     = & (C^2 + C + 1 )\delta
\end{align*}
where the first step follows from triangle inequality, the second step follows from one layer of network is $C$-Lipshitz (see Lemma~\ref{lem:softmv_lipschitz}), and the forth step follows from Lemma~\ref{lem:rank1_multi_head}, and the last step follows from merging the errors.

Therefore for $L$ layers we have
\begin{align*}
    \| \softmv_L(B^0_{L}) - \softmv_L(B^{L}_{L}) \|_{\infty} \leq (C^L + \cdots + C + 1) \delta
\end{align*}
Thus we complete the proof.
\end{proof}

\begin{remark}
We remark that our proof can be straightforwardly generalized to the situation where the Self Attention Network also has MLP layers, similar to Section 3.2 in \cite{dcl21}, by defining $X_{\ell}^{\ell_0} = f(B_{\ell}^{\ell_0})$ where $f$ is the MLP layer. Note that the Lipshitz property of $f$ will appear correspondingly in the final bound.
\end{remark}

\begin{remark}
Recall that above we have $\epsilon_{\ell}=2 \phi_0 \eta (1+H \eta)^{\ell} \in (0,1)$, $\delta= \max_{\ell}2g(2H \epsilon_{\ell})$, and $C = \max_{\ell} 3 \eta (\epsilon_{\ell}^2 + 1)$. There are two different parameter regimes of interest for $\phi_0$ and $\eta$. Case 1, the $\eta$ can be some fixed constant which doesn't depend on $H$ or $L$. In this case, we need $\phi_0$ to exponentially small in $L$, i.e., $\phi_0 \leq 0.5 \eta/(1+H \eta)^{L}$. We have $\delta =O(1)$ and $C \leq 6\eta$. Thus the final bound is $\leq \delta (1+10 C) = O(1) \cdot (1+ O(\eta))$. Case 2. we can choose $\phi_0$ to some fixed constant (bigger than $1$) which doesn't depend on $H$ or $L$. In this case, in order to make $\epsilon_{\ell} \leq 1$, we need to choose $\eta \leq 1/(\phi_0 H L)$. We will have $\delta = O(1)$ and $C\leq 6\eta$. Then the final bound is $\leq O(1)\cdot (1+O(\eta))$.
\end{remark}

\subsection{Lipschitz Property}\label{sec:multiple_layers:lipschitz}
We state a simple application of Lemma~\ref{lem:softmax_a+b_minus_softmax_a_linf}.
\begin{corollary}\label{cor:softmax_a+b_minus_softmax_a_linf}
Let $a, b \in \R^n$. Then, we can show that
\begin{align*}
    \| \soft(a+b) - \soft(a) \|_{\infty} \leq 2(e^{\| b \|_{\infty}}-1)
\end{align*}
\end{corollary}
\begin{proof}
The proof is same as Lemma~\ref{lem:softmax_a+b_minus_softmax_a_linf}.
\end{proof}

\begin{lemma}\label{lem:softmax_lipschitz}
Let $a, b \in \R^n$. If $\| b \|_{\infty} \leq 1$, then we have
\begin{align*}
\| \soft(a+b) - \soft(b) \|_{\infty} \leq 4 \| b \|_{\infty}
\end{align*}
\end{lemma}
\begin{proof}
Note that for $x \in (0,1]$, we know $e^x-1 \leq 2x$.

Thus, we know 
\begin{align*}
\| \soft(a+b) - \soft(b) \|_{\infty} 
\leq & ~ 2(e^{\| b \|_{\infty}}-1) \\
\leq & ~ 4 \| b \|_{\infty}
\end{align*}
where the first step follows from Corollary~\ref{cor:softmax_a+b_minus_softmax_a_linf}, the second step follows from $e^x-1 \leq 2x$.
\end{proof}

\begin{lemma}\label{lem:softmv_lipschitz}
If the following conditions hold
\begin{itemize}
    \item Let $W_q, W_k, W_v$ denote weight matrices.
    \item Let $W = W_q W_k^\top$. 
    \item Let $Y$ satisfy that $\| Y - X \|_{\infty} \leq 2 \| X \|_{\infty}$
    \item $K_1 := 12\| X \|_{\infty} \| W \|_{\infty} $.
    \item $K_2 : = K_1 \| X \|_{\infty} \| W_v \|_{\infty} + \| W_v \|_{\infty} $
\end{itemize}
Then, we can show
\begin{itemize}
\item {\bf Part 1.}
\begin{align*}
    \| \soft(X ) - \soft(Y ) \|_{\infty} \leq K_1 \cdot \| X - Y \|_{\infty}
\end{align*}
\item {\bf Part 2.}
\begin{align*}
    \| \softmv(X) - \softmv(Y) \|_{\infty} \leq K_2 \cdot \| X - Y \|_{\infty}
\end{align*}
\end{itemize}
\end{lemma}
\begin{proof}

{\bf Proof of Part 1.}
We can show
\begin{align*}
    & ~ \| \soft(X ) - \soft(Y) \|_{\infty} \\
    \leq & ~ 4 \| X W X^\top - Y W Y^\top \|_{\infty} \\
    \leq & ~ 4 \| X W X^\top - X W Y^\top \|_{\infty} + 4 \| X W Y^\top - Y W Y^\top \|_{\infty} \\
    \leq & ~ 4 \| X W \|_{\infty} \cdot \| X - Y \|_{\infty} + 4 \| W Y^\top \|_{\infty} \cdot \| X - Y\|_{\infty} \\
    \leq & ~ 4 \cdot ( \| WX \|_{\infty} + \| WY \|_{\infty} )   \cdot \| X - Y \|_{\infty} \\
    \leq & ~ 12 \| W \|_{\infty} \| X \|_{\infty} \| X - Y \|_{\infty}
\end{align*} 
where the first step follows from Lemma~\ref{lem:softmax_lipschitz}, the second step follows triangle inequality, the third step follows from Fact~\ref{fac:norm}, the last step follows from Fact~\ref{fac:norm}.

{\bf Proof of Part 2.}
We can show
\begin{align*}
& ~ \| \softmv(X) - \soft(Y) \|_{\infty} \\
= & ~ \| \soft(X) X W_v - \soft(Y) Y W_v \|_{\infty} \\
\leq & ~ \| \soft(X) X W_v - \soft(Y) X W_v \|_{\infty} + \| \soft(Y) X W_v - \soft(Y) Y W_v \|_{\infty} \\
\leq & ~ \| \soft(X) - \soft(Y) \|_{\infty} \cdot \| X W_v \|_{\infty} + \| (X-Y) W_v \|_{\infty} \\
\leq & ~ K_1 \| X W_v \|_{\infty} \| X- Y \|_{\infty} + \| W_v \|_{\infty} \| X- Y\|_{\infty}
\end{align*}
where the first step follows from definition, the second step follows from triangle inequality, the third step follows from Fact~\ref{fac:norm}, and the last step follows from Part 1 and Fact~\ref{fac:norm}.
\end{proof}

\subsection{Instantiating an Instance for Each Layer Lipschitiz Property}\label{sec:multiple_layers:layer_lipschitz}

\begin{lemma}\label{lem:layer_lipschitz}
If the following conditions hold
\begin{itemize}
    \item Let $X_{\ell}$ denote $\ell$-th layer output
    \item Let $\| W_q \|_{\infty}, \| W_k \|_{\infty} , \| W_v \|_{\infty} \leq \eta$
    \item Let $Y$ satisfy that $\| Y - X_{\ell} \|_{\infty} \leq 2 \| X_{\ell} \|_{\infty}$
    \item $\epsilon_{\ell}:=2\eta \phi_0 (1+H \eta)^{\ell} $.
\end{itemize}
Then, we can show
\begin{itemize}
    \item $\| \softmv(X_{\ell}) -  \softmv(Y)\|_{\infty}
    \leq 3 \eta (\epsilon_{\ell}^2 + 1 ) $
\end{itemize}
\end{lemma}
\begin{proof}
We can show
\begin{align*}
    \| \softmv(X_{\ell}) -  \softmv(Y)\|_{\infty}
    \leq K_2 \cdot \| X - Y \|_{\infty}
\end{align*}
We just need to upper bound $K_2$
\begin{align*}
    K_2 = & ~ K_1 \| X_{\ell} \|_{\infty} \| W_v \|_{\infty} + \| W_v \|_{\infty} \\
    \leq & ~ 12 \| X_{\ell} \|_{\infty}^2 \| W \|_{\infty} \| W_v \|_{\infty} + \| W_v \|_{\infty}  \\
    \leq & ~ 12 \| X_{\ell} \|_{\infty}^2 \eta^3+ \eta \\
    \leq & ~ 12 ( \phi_0 \cdot (1+H \eta)  )^2 \eta^3 + \eta \\
    = & ~ 3 \eta( \epsilon_{\ell}^2 + 1 )
\end{align*}
where the first step follows from the definition of $K_2$, the forth step follows from Lemma~\ref{lem:layer_norm}, and the fifth step follows from the definition of $\epsilon_{\ell}$.
\end{proof}

\subsection{Each Layer Norm is not Increasing Much}\label{sec:multiple_layers:layer_norm}
\begin{lemma}\label{lem:layer_norm}
If the following conditions hold
\begin{itemize}
    \item Let $X_0$ denote the input of first layer of neural network, and satisfy $\| X_0 \|_{\infty} \leq \phi_0$
    \item For $\ell \in [L]$, we use $X_{\ell}$ to denote the $\ell$-th layer output
    \item Let $\| W_v \|_{\infty} \leq \eta$
\end{itemize}
Then, we can show
\begin{itemize}
    \item Part 1. For any $\ell$, $\| X_{\ell+1} \|_{\infty} \leq \| X_{\ell} \|_{\infty} \cdot (1+H \eta)$
    \item Part 2. For any $\ell$, $\| X_{\ell} \|_{\infty} \leq \phi_0 \cdot (1+H \eta)^{\ell}$
\end{itemize}
\end{lemma}
\begin{proof}

{\bf Proof of Part 1.}

For any $\ell$, we have
\begin{align*}
    \| X_{\ell+1} \|_{\infty}
    = & ~ \| X_{\ell} + \sum_{i=1}^H \softmv_i(X_{\ell}) \|_{\infty} \\
    \leq & ~ \| X_{\ell} \|_{\infty} + H \cdot \| \softmv_i(X_{\ell}) \|_{\infty} \\
    = & ~ \| X_{\ell} \|_{\infty} + H \cdot \| \soft_i(X_{\ell}) X_{\ell} W_{v,i} \|_{\infty} \\
    \leq & ~ \| X_{\ell} \|_{\infty} + H \cdot \| \soft_i(X_{\ell}) \|_{\infty} \cdot \| X_{\ell}\|_{\infty} \cdot \| W_{v,i} \|_{\infty} \\
    \leq & ~ \| X_{\ell} \|_{\infty} (1 + H \eta)
\end{align*}
where the first step follows from definition of $X_1$, the second step follows from triangle inequality, the third step follows from definition of $\softmv$, the forth step follows from Fact~\ref{fac:norm}, and the last step follows from $\| \soft() \|_{\infty} \leq 1$ and $\| W_{v,i} \|_{\infty} \leq 1$.  (Here $W_{v,i} \in \R^{n \times d}$ denotes the weight matrix, $W_v$, for the $i$-th head.)

{\bf Proof of Part 2.}

We can show
\begin{align*}
    \| X_{\ell} \|_{\infty} 
    \leq & ~ \| X_{\ell-1} \|_{\infty} (1+ H \eta) \\
    \leq & ~ \cdots \\
    \leq & ~ \| X_0 \|_{\infty} (1+H \eta)^{\ell} \\
    \leq & ~ \phi_0 \cdot (1+H \eta)^{\ell}
\end{align*}
where the first step follows from Part 1, the third step follows from recursively applying Part 1, and the last step follows from $\| X_0 \|_{\infty} \leq \phi_0$.

Therefore, we complete the proof.
\end{proof}

\section{Limitations}\label{sec:limitation}

There are several limitations in our work, which we leave as open problems for future research. Currently our result mainly focus on $\ell_{\infty}$. We believe it is also worthwhile to consider the $\ell_2$ norm or $\ell_1$ norm in future work, as they may give more meaningful bounds in certain parameter regimes.

%This work is entirely theoretical and does not include empirical experiments to validate the findings.

\section{Broader Impact}\label{sec:impact}

Our results offer new theoretical insights into the expressiveness of attention mechanisms in transformers. These findings may guide the future design of large language models toward more expressive architectures. We do not foresee any potential negative societal impacts from this work.

\ifdefined\isarxiv
\newpage
\bibliographystyle{alpha}
\bibliography{ref}
\else

\fi

\end{document}